\documentclass{article}
\usepackage{arxiv}
\usepackage{natbib}
\usepackage{booktabs}
\usepackage{graphicx}
\usepackage{amsmath}
\usepackage{multirow}
\usepackage[linesnumbered, ruled, vlined]{algorithm2e}
\usepackage{algorithmic}
\usepackage{pifont}

\AtBeginDocument{%
  \providecommand\BibTeX{{%
    \normalfont B\kern-0.5em{\scshape i\kern-0.25em b}\kern-0.8em\TeX}}}

\begin{document}
\title{Out-of-Distribution Detection in Dermatology using Input Perturbation and Subset Scanning}

\author{%
  Hannah Kim \\
  Duke University\\
  Durham, NC, USA\\
  % Cranberry-Lemon University\\
  % Pittsburgh, PA 15213 \\
  % \texttt{hippo@cs.cranberry-lemon.edu} \\
  \And
  Girmaw Abebe Tadesse \\
  IBM Research\\
  Nairobi, Kenya\\
  \And
  Celia Cintas\\
  IBM Research\\
  Nairobi, Kenya\\
  \And
  Skyler Speakman\\
  IBM Research\\
  Nairobi, Kenya\\
  \AND
  Kush Varshney\\
  IBM Research\\
  Yorktown Heights, NY, USA\\
}
% \author{Anom1}
% %\authornote{Both authors contributed equally to this research.}
% \affiliation{%
%   \institution{WW}
%   %\streetaddress{P.O. Box 1212}
%   \city{WW}
%   %\state{Ohio}
%   \country{WW}
%   %\postcode{43017-6221}
% }
% \email{email}
% \orcid{XXX}

% \author{Anom2}
% %\authornotemark[1]
% \email{email}
% \affiliation{%
%   \institution{WW}
%   %\streetaddress{P.O. Box 1212}
%   \city{WW}
%   %\state{Ohio}
%   \country{WW}
%   %\postcode{43017-6221}
% }
% 
% \author{Anom3}
% \affiliation{%
%   \institution{WW}
%   %\streetaddress{1 Th{\o}rv{\"a}ld Circle}
%   \city{WW}
%   \country{WW}}
% \email{email}

% \author{Anom4}
% \affiliation{%
%   \institution{WW}
%   \city{WW}
%   \country{WW}
% }

% \author{Anom4}
% \affiliation{%
%  \institution{XX}
%  %\streetaddress{Rono-Hills}
%  \city{XX}
%  %\state{Arunachal Pradesh}
%  \country{ZZZ}}

% \author{Anom5}
% \affiliation{%
%   \institution{WWW}
%   %\streetaddress{30 Shuangqing Rd}
%   \city{ZZZZ}
%   %\state{Beijing Shi}
%   \country{OOO}}
\maketitle
\begin{abstract}
Recent advances in deep learning have led to breakthroughs in the development of  automated skin disease classification. As we observe an increasing interest in these models in the dermatology space, it is crucial to address aspects such as the robustness towards input data distribution shifts. 
Current skin disease models could make incorrect inferences for test samples from different hardware devices and clinical settings or unknown disease samples, which are out-of-distribution (OOD) from the training samples.To this end, we propose a simple yet effective approach that detect these OOD samples prior to making any decision. The detection is performed via scanning in the latent space representation (e.g., activations of the inner layers of any pre-trained skin disease classifier). The input samples could also perturbed to maximise divergence of OOD samples.
We validate our ODD detection approach in two use cases: 1) identify samples collected from different protocols, and 2) detect samples from unknown disease classes. Additionally, we evaluate the performance of the proposed approach and compare it with other state-of-the-art methods. Furthermore, data-driven dermatology applications may deepen the disparity in clinical care across racial and ethnic groups since most datasets are reported to suffer from bias in skin tone distribution. Therefore, we also evaluate the fairness of these OOD detection methods across different skin tones.  Our experiments resulted in competitive performance across multiple datasets in detecting OOD samples, which could be used (in the future) to design more effective transfer learning techniques prior to inferring on these samples. 

% \vspace{-5mm}
\end{abstract}

%   <concept>
%       <concept_id>10010147.10010257.10010293.10010294</concept_id>
%       <concept_desc>Computing methodologies~Neural networks</concept_desc>
%       <concept_significance>500</concept_significance>
%       </concept>
%   <concept>
%       <concept_id>10010147.10010257.10010258.10010260.10010229</concept_id>
%       <concept_desc>Computing methodologies~Anomaly detection</concept_desc>
%       <concept_significance>500</concept_significance>
%       </concept>
%   <concept>
%       <concept_id>10010147.10010178.10010224.10010225.10011295</concept_id>
%       <concept_desc>Computing methodologies~Scene anomaly detection</concept_desc>
%       <concept_significance>300</concept_significance>
%       </concept>
%  </ccs2012>

% \ccsdesc[500]{Computing methodologies~Neural networks}
% \ccsdesc[500]{Computing methodologies~Anomaly detection}
% \ccsdesc[300]{Computing methodologies~Scene anomaly detection}

  %\includegraphics[width=\textwidth]{sampleteaser}
  %\caption{Seattle Mariners at Spring Training, 2010.}
  %\Description{Enjoying the baseball game from the third-base
  %seats. Ichiro Suzuki preparing to bat.}
  %\label{fig:teaser}

\section{Introduction}
% \vspace{-2mm}
Skin disease remains a global health challenge, with skin cancer being the most common cancer worldwide~\citep{ISIC-2019}. Following the recent success of deep learning (DL) in various computer vision problems (partly due to its automated feature encoding capability), convolutional neural networks (CNNs)~\citep{densenet}
%,mobilenet} 
have been employed for skin disease classification tasks. 
As we observe increasing interest in DL in applying dermatology~\citep{esteva2017dermatologist,gomolin2020artificial}%,chan2020machine}
, it is imperative to address transparency, robustness, and fairness of these solutions~\citep{adamson2018machine,qayyum2020secure}. %,grote2020ethics}. 
While many existing deep learning techniques~\citep{mahbod,gessert,ahmed} %,gessert3,pacheco2} 
achieve high performance on publicly available datasets~\citep{ISIC-2019,HAM10000,BCN20000,sd-198}, they utilize ensembles of multiple models aimed at maximising performance with limited consideration to shifts in the input data~\citep{ahmed,gessert3,zhang}, which might result in incorrectly classifying  new samples as one of the training classes (with high confidence) though these samples might be from previously unknown or new classes. 

Thus, it is necessary to detect out-of-distribution (OOD) samples prior to making decisions in order to  achieve principled transfer of knowledge from in-distribution training samples to OOD test samples, thereby extending the usability of the models to previously unseen scenarios.
% ~\cite{Karimi,cao,Adler}. 
% Note known disease types refer to those for which the model is trained on.
Furthermore,  OOD detectors and other DL solutions need to guarantee equivalent detection capability across sub-populations. Particularly in dermatology, bias in representations of skin tones in academic materials~\citep{Mcfarling_2020} and clinical care \citep{rabin_2020} is becoming a primary concern. %\hannah{
For instance, New York Times reports major disparities in dermatology when treating skin of color~\citep{rabin_2020} as common conditions often manifest differently on dark skin, and physicians are trained mostly to diagnose them on light skin. STAT~\citep{Mcfarling_2020} also reported that lack of darker skin tones in dermatology academic materials adversely affects the quality of care for patients of color. Alarmingly, the growing practice of using artificial intelligence to aid the diagnosis of skin diseases will further deepen the divide in patient care because of the machine learning algorithms, which are trained with such imbalanced datasets~\citep{ISIC-2018,ISIC-2019,HAM10000,BCN20000,sd-198} (with overwhelming majority of samples with light skin tones). This is supported by the work of Kinyanjui~\textit{et al.}~\citep{skinTone}, which use Individual Typology Angle (ITA) to approximate skin tones in various publicly available skin disease datasets~\citep{ISIC-2019,HAM10000,BCN20000,sd-198} and show that these datasets heavily under-represent darker skin tones. As a result, we also validated the performance of 
% We further extend this work to validate the impact of variations in skin tones in detecting OOD samples.%}

%\hannah{
To this end, we propose a simple yet effective approach that scans over the activations of the inner layers of any pre-trained skin disease classifier to detect OOD samples. We additionally perturb the input data beforehand with our proposed ODIN$_{low}$, a modification of ODIN~\citep{liang}, which improve OOD detection performance in earlier layers of the network. In our framework, we define two different OOD use cases: \textit{protocol variations} (e.g., different hardware devices, lighting settings and not compliant with clinical protocol);  and \textit{unknown disease types} (e.g., samples from new disease type that was not observed during training). 
% We validate our approach with two types of OOD samples, those that are collected from different protocol or equation and those that are from unknown disease classes. 
Without requiring any prior knowledge of the OOD samples,  the proposed approach out-performed  existing OOD detectors, softmax score~\citep{hendrycks} and ODIN~\citep{liang}, for OOD samples with different validation protocols, and competitive performance is achieved in detecting samples with unknown disease types. 
We further explore how our proposed and existing OOD detectors perform across skin tones to evaluate fairness. We show that the current OOD detectors show higher performance in detecting darker skin tones as OOD samples than those of lighter skin tones, which is likely impacted by the imbalanced training skin datasets that heavily lack samples of dark skin tones.%}

%We consider two types of OOD samples: 1) samples that are collected from different collection environments or protocols compared to those of the training data, and 2) samples of unknown diseases that the CNNs have not seen before. 

% For these reasons, it is imperative to make sure that the detection mechanism will be able to perform equally across different skin tones.

Generally, our main contributions are highlighted as follows: 1) We propose a weakly-supervised approach based on subset scanning over the activations of the inner layers of a pre-trained skin disease classifier to detect OOD samples across two use cases: detection of OOD samples from different collection protocol and those from unknown disease classes; 2)%\hannah{
We propose perturb input images with ODIN$_{low}$ noise, %a modification of ODIN noise, 
for improved OOD detection performance;%} 
3) We evaluate our methods against existing OOD detectors: Softmax Score~\citep{hendrycks} and ODIN~\citep{liang}; Furthermore, we evaluate the fairness  of the proposed approach and existing methods in their detection performance across skin tones. 

\begin{table*}[!htbp]
    \centering
   \resizebox{1\linewidth}{!}{
    \begin{tabular}{c|ccccccc}
        \toprule
       % \multicolumn{2}{c|}{Methods} 
        & Ensemble & \begin{tabular}{c}Test Data\\Augmentation\end{tabular} & \begin{tabular}{c}OOD Detection\\ Post-Training\end{tabular} & \begin{tabular}{c}New Protocol \\ Detection\end{tabular}& \begin{tabular}{c}New Disease \\ Detection\end{tabular}& \begin{tabular}{c}Algorithmic \\ Fairness \end{tabular}\\ \midrule
        \citep{ahmed} &\ding{51} & \ding{51} & \ding{55} & \ding{55}&\ding{51}  &\ding{55}\\
        % Zhang\textit{et al.}~
        \citep{zhang} &\ding{51}& \ding{55} & \ding{55}& \ding{55} &\ding{51}  &\ding{55}\\
        % Gessert \textit{et al.}~
        \citep{gessert3}  &\ding{51} &\ding{51}&\ding{55} & \ding{55}& \ding{51} &\ding{55} \\
        % Bagchi \textit{et al.}~
        \citep{bagchi}  & \ding{55} & \ding{55} & \ding{55} & \ding{55}&\ding{51}  &\ding{55}\\
        % \midrule
        % Pacheco \textit{et al.}~
       % \multirow{3}{*}{Post-training}&
       \citep{pacheco2} & \ding{51}&\ding{55}&\ding{51}& \ding{55} &\ding{51}  &\ding{55}\\
        % Combalia \textit{et al.}~
        \citep{combalia} & \ding{55} & \ding{51} & \ding{51} & \ding{55} &\ding{51}  &\ding{55}\\
        % Pacheco \textit{et al.}~
        \citep{pacheco} & \ding{55} &\ding{55} & \ding{51} & \ding{51} & \ding{51} &\ding{55}\\ \midrule
         Ours & \ding{55}  &\ding{55}&\ding{51}&\ding{51}&\ding{51}&\ding{51} \\ 
        \bottomrule
    \end{tabular}
    }
    \caption{Summary of the state-of-the-art OOD sample detection in skin disease classification task, and the differentiation of our proposed approach. 
    }
    \label{tab:ISIC_work}
    % \vspace{-9mm}
\end{table*}

% \vspace{-0.3cm}
\section{Related Work} \label{relatedworks}

Our review of existing OOD detection methods is grouped into \textit{pre-training}~\citep{ahmed,bagchi,gessert3,zhang} and \textit{post-training}~\citep{combalia,pacheco2,pacheco}, based on where the detection step is applied. % relative to training. 

\textbf{Pre-training OOD detection} approaches have prior knowledge of the OOD samples and incorporate it during their training phases. Many of these approaches utilize ensembles of existing CNNs (and their variants) to detect OOD samples~\citep{ahmed,gessert3,zhang}. Ahmed~\textit{et al.}~\citep{ahmed} applied one-class learning using deep neural network features where one-class samples were iteratively discarded as OOD samples in a one-vs-all cross-validation strategy, and the OOD samples were detected by taking the prediction average of all the models. Gessert~\textit{et al.}~\citep{gessert3} utilized an additional dataset of skin lesions as OOD samples to train their ensemble of CNNs to detect OODs. Zhang~\textit{et al.}~\citep{zhang} employed an ensemble DenseNet-based CNNs consisting of both multi-class and binary classifiers to detect OOD samples. Bagchi~\textit{et al.}~\citep{bagchi} proposed \textit{Class Specific - Known vs. Simulated Unknown} to detect OOD samples.

\textbf{Post-training OOD detection} approaches do not require any prior knowledge of the OOD samples during training~\citep{combalia,pacheco2,pacheco}. Pacheco~\textit{et al.}~\citep{pacheco2} detected OOD samples using \textit{Shannon entropy}~\citep{Shannon} and \textit{cosine similarity} metrics on their CNN's probability outputs. Instead, Combalia~\textit{et al.}~\citep{combalia} detected OOD samples using \textit{Monte-Carlo Dropout}~\citep{gal} and test data augmentation
% ~\cite{ayhan,wang} 
 to estimate uncertainty such as entropy and variance in their network predictions. Pacheco~\textit{et al.}~\citep{pacheco} extended Gram-OOD~\citep{sastry} with layer-specific normalization of Gram Matrix values to detect OOD samples.

Table~\ref{tab:ISIC_work} summarizes notable OOD detection studies in dermatology. The majority of these studies employ pre-training approaches using ensembles of CNNs, which result in model complexity and impracticality due to their need of prior knowledge of OOD samples. Test data augmentation is also less plausible to domain experts as it might partially re-synthesize the samples. In this work, we propose a simple, post-training OOD detector that can be applied to any single pre-trained network without any test data augmentation nor prior knowledge of the OOD samples.

\begin{figure*}[hbtp!]
    \centering
    % \toprule
    \includegraphics[width=0.9\linewidth]{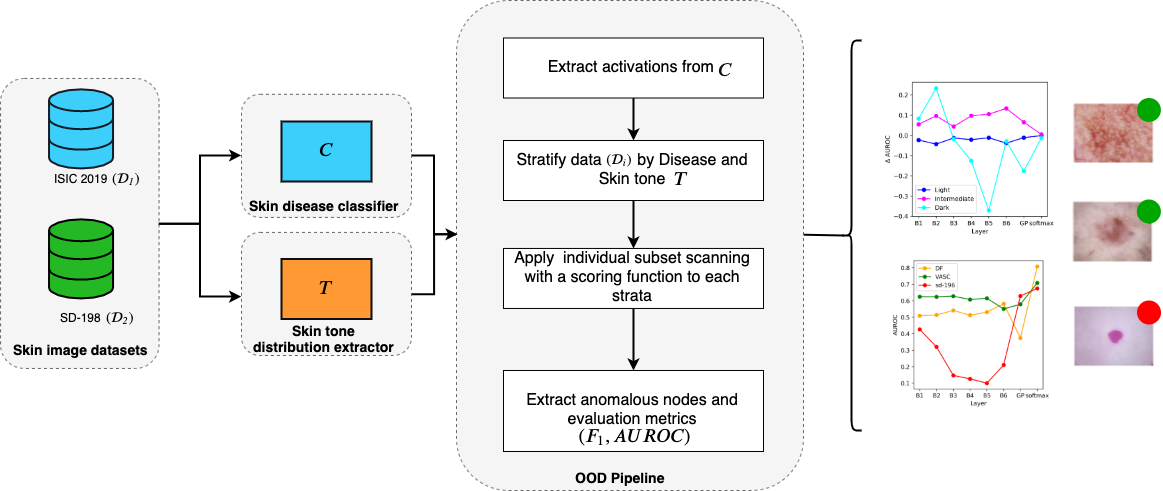} 
    % \bottomrule
    \caption{Block diagram of the proposed approach.  $C$: a trained model for skin disease classification over mentioned datasets ($\mathcal{D}_1$, $\mathcal{D}_2$); $T$: a skin tone extractor.}
    \label{fig:approach}
    % \vspace{-1mm}
\end{figure*}

\section{Proposed Framework}\label{proposed}
% \vspace{-2mm}

We propose a weakly-supervised OOD detection method to identify skin images collected in different validation protocols and derived from unknown skin disease types, based on subset scanning~\citep{cintas2020detecting} and ODIN~\citep{liang}. Subset scanning treats the OOD detection problem as a search for the {\em most anomalous} subset of observations in the activation space of any pre-trained classifier. This exponentially large search space is efficiently explored by exploiting mathematical properties of our measure of anomalousness~\citep{neill-ltss-2012}. Our solution can be applied to any off-the-shelf skin disease classifier. Additionally, we evaluate algorithmic fairness of the proposed and existing OOD detectors across skin tones.
The overview of the proposed approach is shown in Fig.~\ref{fig:approach}. 
Given a set of skin datasets $D$ and a pre-trained skin disease classifier $C$ as an input; first, we stratify each dataset through a skin tone distribution extractor $T$ for evaluation purposes. Then, we apply subset scanning across each layer of the classifier $C$ and compute the subset score for the unknown disease use case. To detect protocol variations, we first perturb the input data for the best performing results. 
% Lastly, for all cases we compute metrics for further evaluation. 
In the following sections, we describe the details of the proposed approach.
% for OOD detection.

%  \vspace{-1cm}
\subsection{Subset scanning for out-of-distribution sample detection}
Given a pre-trained network $C$ for skin disease classification, we apply subset scanning~\citep{cintas2020detecting} on the activations in the intermediate layers of the network $C$ to detect a subset ($S$) of OOD samples (see Algorithm \ref{alg:algo_disjdecomp}). Subset scanning searches for the most anomalous subset $S^*=\arg\max_{S}F(S)$  in each layer, where the anomalousness is quantified by a scoring function $F(\cdot)$, such as a log-likelihood ratio statistic. When searching for this subset, an exhaustive search across all possible subsets is computationally infeasible as the number of subsets ($2^N$) increases exponentially with the number of nodes ($N$) in a layer. Instead, we utilize a scoring function that satisfies the Linear Time Subset Scanning (LTSS)~\citep{neill-ltss-2012} property, which enables  efficient maximization over all subsets of data. This LTSS property guarantees that the highest-scoring subset of nodes in a layer are identified within $N$ searches instead of $2^N$ searches. Following the literature on pattern detection~\citep{FGSS}, we utilize non-parametric scan statistics (NPSS)~\citep{FGSS} as our scoring function as it satisfies LTSS property and makes minimal assumptions on the underlying distribution of node activations. 

We apply subset scanning on set of layers $C_Y$ of our pre-trained network $C$. For each layer $C_y \in C_Y$, we form a distribution of expected activations at each node using the known in-distribution (ID) samples $X_z$, which were used during training and can also be referred as background images. Comparing this expected distribution to the node activations of each test or evaluation sample $X_i$, we can obtain p-values $p_{ij}$ for each $i^{th}$ test sample and $j^{th}$ node of layer $C_y$. We can then quantify the anomalousness of the p-values by finding the subset of nodes that maximize divergence of the test sample activations from the expected. This yields $|C_Y|$ anomalous scores $S^\ast_{(C_y)}$ for each test sample. We expect OOD samples to yield higher anomalous scores $S$ than ID samples, and we detect OOD samples with simple thresholding. Note that the OOD detection is performed in an unsupervised fashion without any prior knowledge of the OOD samples.

% \vspace{1cm}
\begin{algorithm}[hbtp!]
\caption{Pseudo-code for the proposed new protocol (OOD) detection.}\label{alg:algo_disjdecomp}
%\begin{algorithmic}[1]
\SetKwFunction{TrainSkinDiseaseClassifier}{TrainSkinDiseaseClassifier}
\SetKwFunction{ExtractActivation}{ExtractActivation}
% \SetKwFunction{ExtractNonDiseasedRegions}{ExtractNonDiseasedRegions}
\SetKwFunction{PredictITASkinTone}{PredictITASkinTone}
\SetKwFunction{ComputeDetectionPerformance}{ComputeDetectionPerformance}
\SetKwFunction{StratifyPerSkinTone}{StratifyPerSkinTone}
\SetKwFunction{AddODINNoise}{AddODINNoise}
\SetKwFunction{SortAscending}{SortAscending}
\SetKwFunction{NPSS}{NPSS}

\SetKwInOut{Input}{input}
\SetKwInOut{Output}{output}
\Input{Background Image: $X_z \in D^{H_0}$, Evaluation Image: $X_i$, training dataset: $D_{train}$, $\alpha_{\text{max}}$.}
\Output{$AUROC$, $F_1$, $AUROC^{t}$, and $F_1^{t}$  \CommentSty{ for $X_i$}} %\textcolor{red}{also mention the subset of nodes not just the score?}
%\Parameter{A parameter for the algorithm}
\BlankLine
$ C \leftarrow$ \TrainSkinDiseaseClassifier($D_{train}$)\;
$ C_Y \leftarrow$ Set of layers in $C$\;
% $X_i^{nd} \leftarrow$ \ExtractNonDiseasedRegions($X_i$)\;
$X_i^{t} \leftarrow$ \PredictITASkinTone($X_i$)\;

% \hannah{Might get rid of below if statement depending on the final story} \
%\If{$X_i$ collected from a different protocol or equipment:}{
$\hat{X}_z \leftarrow$ \AddODINNoise($X_z$)\;
$\hat{X}_i \leftarrow$ \AddODINNoise($X_i$) \; %}
\For{$C_y$ in $C_Y$}{
%\item For each image $X_i$ and each Node $O_j$, in both 
%validation and evaluation datasets, compute the activation $A_{ij}$ given the 
%network.
%\For{$z\leftarrow 0$ \KwTo $M$}{
\For{$j\leftarrow 0$ \KwTo $|C_y|$}{
 $A^{H_0}_{zj} \leftarrow $ \ExtractActivation($C_{y}$, $\hat{X}_z$)\;
 
 $A_{ij} \leftarrow $ \ExtractActivation($C_{y}$,$\hat{X}_i$)\;}

%\For {$j\leftarrow 0$ \KwTo $J$}{
%}
%\For {$j\leftarrow 0$ \KwTo $J$}{
$p_{ij} = \frac{\sum_{X_z \in D^{H_0}} I(A_{zj} >= A_{ij} ) + 1}{M+1}$\;
%}
$p^{\ast}_{ij} = \{y < \alpha_{\text{max}} \: \forall \: y \subseteq p_{ij} \}$\;
$p^s_{ij} \leftarrow $ \SortAscending($p^{\ast}_{ij}$)\;
\For {$k\leftarrow 1$ \KwTo $|C_y|$}{
$S_{(k)} = \{p_y \subseteq p^s_{ij} \forall y \in \{1, \ldots, k\} \}$\;
$\alpha_{k} = max(S_{(k)})$\;
$F(S_{(k)}) \leftarrow $ \NPSS($\alpha_{k}$, k, k)\; 
}
$k^{\ast}_{(C_y)} \leftarrow  \arg\max F(S_{(k)})$\;
$\alpha^{\ast}_{(C_y)} = \alpha_{k^\ast_{(C_y)}}$\;
$S^\ast_{(C_y)} = S_{(k^\ast_{(C_y)})}$\;
}%}
$AUROC$, $F_1$ = \ComputeDetectionPerformance($\sum_{C_y}{S^\ast_{(C_y)}}$)\;
$AUROC^{t}$, $F_1^{t}$ = \StratifyPerSkinTone{$X_i^{t}$, $AUROC$, $F_1$}\;
\Return $AUROC$, $F_1$, $AUROC^{t}$, and $F_1^{t}$ 
%\end{algorithmic}
\end{algorithm}

\subsection{ODIN and ODIN$_{low}$ Perturbations}
We have also evaluated the impact of adding small perturbations, prior to subset scanning, to each test sample following ODIN~\citep{liang} for enhanced OOD. ODIN involves two steps, input pre-processing and temperature scaling. In the first step, $X_i$ is perturbed by adding a small perturbation computed by back-propagating the gradient of the training loss with respect to $X_i$ and weighted by parameter $\epsilon$. This pre-processed $X_i$ is then fed into the neural network and temperature scaling  with parameter $\tau$ is applied in the final softmax layer $C_{s}$. The two hyperparemters, $\epsilon$ and $\tau$, are chosen so that the OOD detection performance of softmax score~\citep{hendrycks}, the maximum value of the softmax layer output, is optimized. We further modified ODIN and propose ODIN$_{low}$ with parameters $\tau_{low}$ and $\epsilon_{low}$ that leads to the lowest softmax score performance. %\hannah{
As subset scanning is applied not only on the softmax layer but also on the the inner layers of the network, we show that ODIN$_{low}$ helps improve OOD detection in the earlier layers of the network.%}

\subsection{Algorithmic Fairness of OOD detectors across skin tone} \label{fairness}
We further evaluate algorithmic fairness of our proposed OOD dectector across skin tones, estimated by adopting an existing framework~\citep{skinTone}. To this end, the non-diseased regions of a given skin image are segmented using Mask R-CNN~\citep{maskrcnn}, and individual typology angle (ITA) values are computed as $ITA = \arctan\left(\frac{L_{\mu}-50}{b_{\mu}}\right) \times\frac{180^{\circ}}{\pi}$, where $L_{\mu}$ and $b_{\mu}$ are the average of luminance and yellow values of non-diseased pixels in CIELab-space.   ITA values are used to stratify the samples into three Fitzpatrick skin tone categories, Light, Intermediate, and Dark, as shown in Table~\ref{tab:ITA_categories}.

\setlength{\tabcolsep}{25pt}
\begin{table}[htbp!]
    \centering
    \begin{tabular}{c|c} \toprule
        ITA Range & Skin Tone Category \\ \midrule
        $ITA > 41^{\circ}$ & Light \\
        $28^{\circ} < ITA \leq 41^{\circ}$ & Intermediate \\
        $ITA \leq 28^{\circ}$ & Dark \\ \bottomrule
    \end{tabular}
    \caption{Summary of Fitzpatrick skin tone categorization of computed $ITA$ values.}
    \label{tab:ITA_categories}
\end{table}

% \vspace{-0.35cm}
\section{Datasets}\label{datasets}
% \vspace{-2mm}
We validate the proposed frame work using two datasets: ISIC 2019~\citep{ISIC-2019,HAM10000,BCN20000} for samples of unknown diseases; and SD-198~\citep{sd-198} for samples from unknown collection protocols. We stratify the samples from both datasets based on skin-tones to observe the impact of various OOD methods across the population spectrum (see Figure~\ref{fig:eximgs_skin_tone}).

\subsection{ISIC 2019}
ISIC 2019~\citep{ISIC-2019,HAM10000,BCN20000} dataset is an extension of ISIC 2018 and merges HAM10000~\citep{HAM10000}, BCN20000~\citep{BCN20000}, and MSK~\citep{ISIC-2019} datasets. It consists of $25,331$ dermoscopic images among eight diagnostic categories: \textit{Melanoma, Melanocytic nevus, Basal cell carcinoma, Actinic keratosis, Benign keratosis, Dermatofibroma, Vascular lesion}, and \textit{Squamous cell carcinoma}. As its test set is not available publicly, we set aside \textit{Dermatofibroma} (DF) and \textit{Vascular lesion} (VASC) samples during training, and utilize them during the test time as OOD samples of unknown diseases. These two classes are chosen as they contain the least number of samples in the dataset. First row of Figure \ref{fig:eximgs_skin_tone} show example images of this dataset for each of the three skin tone categories we consider in this work. 

\subsection{SD-198}
\begin{figure}[]
    \centering
    % \resizebox{\linewidth}{!}{
    % \begin{tabular}{cccccc|cccccc}
    % \toprule
    %     \multicolumn{6}{c|}{ISIC 2019} & \multicolumn{6}{c}{SD-198} \\
    %     Light & Intermediate & Dark &&&&&&& Light & Intermediate & Dark\\ \midrule
    %     \includegraphics[width = 0.11\textwidth]{images/VASC_light2.png} & \includegraphics[width = 0.11\textwidth]{images/DF_med2.png} & 
    %     \includegraphics[width = 0.11\textwidth]{images/DF_dark2.png} & & & & & & &
    %     \includegraphics[width = 0.11\textwidth]{images/light_3.png} & \includegraphics[width = 0.11\textwidth]{images/intermediate_4.png} & 
    %     \includegraphics[width = 0.11\textwidth]{images/dark_1.png}\\
    %     \bottomrule
    % \end{tabular}
    % }
    
    \begin{tabular}{ccc}
    \toprule
        % \multicolumn{3}{c}{ISIC 2019} \\
        % Light & Intermediate & Dark \\ \midrule
        \includegraphics[width = 0.14\textwidth]{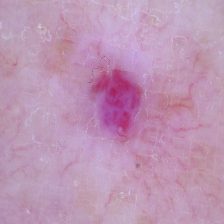} & \includegraphics[width = 0.14\textwidth]{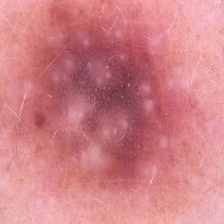} &
        \includegraphics[width = 0.14\textwidth]{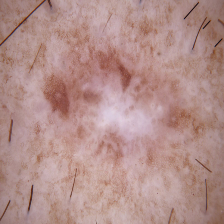} \\ 
        % \bottomrule
        % \multicolumn{3}{c}{SD-198} \\
        \includegraphics[width = 0.14\textwidth]{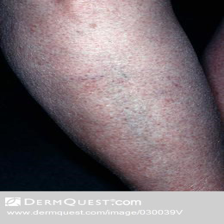} & \includegraphics[width = 0.14\textwidth]{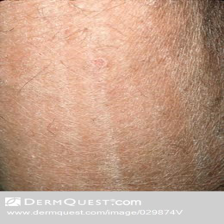} & 
        \includegraphics[width = 0.14\textwidth]{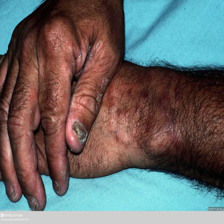}\\ 
        % \midrule
        Light & Intermediate & Dark \\
        \bottomrule
    \end{tabular}
    \caption{Example images from ISIC 2019~\citep{ISIC-2019} (top) and SD-198~\citep{sd-198} (bottom) stratified into three skin tone categories: Light, Intermediate, and Dark.}
    \label{fig:eximgs_skin_tone}
    % \vspace{-6mm}
\end{figure}

SD-198~\citep{sd-198} dataset contains $198$ different diseases from different types of eczema, acne and various cancerous conditions, totalling $6,584$ images. The images are collected via various devices, mostly digital cameras and mobile phones with higher levels of noise and varying illumination. We use this dataset for OOD samples that are collected from unknown protocols. % It contains $198$ different diseases from different types of eczema, acne and various cancerous conditions, totalling $6,584$ images. % We use this dataset during testing time as our OOD samples that are collected from a different equipment or protocol from the training dataset, ISIC 2019~\cite{ISIC-2019,HAM10000,BCN20000}.
We show some example images of the dataset in the second row of Figure \ref{fig:eximgs_skin_tone} that are stratified into three skin-tone categories, Light, Intermediate, and Dark. % SD-198-P~\cite{sd-198-p} additionally includes high-level position information to SD-198 dataset.

\setlength{\tabcolsep}{3pt}

\section{Experimental setup} \label{exps}
%\vspace{-2mm}
% \subsubsection{Experimental setup} \label{setup} 
\subsection{Skin disease model setup} We adopt DenseNet-121~\citep{densenet} pre-trained on ImageNet~\citep{imagenet} for the skin disease classification task and fine-tune it on ISIC 2019~\citep{ISIC-2019}. To accommodate for the change in number of classes for the skin disease classification task, we resize the last four fully connected layers of DenseNet to $512$, $256$, $128$, and $7$ nodes followed by a SoftMax with $7$ nodes for the seven skin disease classes. We use Adam~\citep{adam} optimization with a learning rate of $1e^{-4}$ and a batch size of $40$. To address the class imbalance problem, we employ weighted cross-entropy loss. The implementation is done with the Python 3.6~\citep{harris2020array} and TensorFlow 1.14~\citep{TF}. 
%We train on \hannah{Tensorflow} Adam~\cite{adam} with a learning rate of $1e^{-4}$ using a batch size of 40. To address the class imbalance problem, we employ weighted cross-entropy loss.
% To validate unknown disease We experiment with two types of OOD samples, one for samples of unknown diseases, 
To validate detection of unknown disease samples, we use DF and VASC classes from ISIC-2019, consisting of $253$ and $225$ samples, respectively. Similarly, for samples with different collection protocols, we extract $10$ sets of $260$ samples from SD-198 and report their aggregate performance. 
%  For in-distribution (ID) samples, we use 248 samples of ISIC-2019 dataset that are unseen during the training of our adapted DenseNet-121. 
% Using these OOD samples, we employ subset scanning on the node activations of inner layers of our trained DenseNet-121~\cite{densenet}. 
\subsection{Subset scanning setup} We apply subset scanning across eight layers $C_Y$ consisting of six convolutional layers $(C_{conv_1},...,C_{conv_6})$, global pooling layer $(C_{gp})$, and softmax layer $(C_{s})$.
% , for each of the ID and OOD samples.
% For an improved OOD detection, we add noise, ODIN and ODIN$_{low}$ to the samples before applying subset scanning. 
% ODIN~\cite{liang} involves two hyper-parameters, $\epsilon$ and $\tau$, which are chosen based on the softmax score or the maximum value of the softmax layer output as detailed above. 
For ODIN~\citep{liang}, we use temperature scaling parameter $\tau = 10$ and perturbation magnitude $\epsilon = 0$ (optimized on ISIC-2019) for SD-198 samples and $\tau = 5$ and $\epsilon = 0.0002$ (optimized on SD-198) for ISIC-2019 samples. For ODIN$_{low}$, we use $\tau_{low} = 2$ and $\epsilon_{low} = 0.2$, which leads to AUROC equal to 0.5 for Softmax Score for both OOD use cases. We employ Area Under Receiver Operating Characteristic Curve (AUROC) and maximum $F_1$-score ($F_1$) as our metrics to evaluate the OOD detection performance.

\section{Results}
In this section, we show the result of proposed OOD detector with subset scanning and ODIN as detailed in Section \ref{proposed}. We first compare our result of OOD detection to Softmax Score~\citep{hendrycks} and ODIN~\citep{liang} in Tables \ref{tab:lr_bin_sd} for OOD samples with different collection protocol and in \ref{tab:lr_bin_isic} for OOD samples with  unknown disease types. We further stratify OOD samples based on skin tone for these approaches and report their performance in Table \ref{tab:lr_ita_bin}. We show in Figure \ref{fig:ss_plot} the detection performance of our proposed method on individual layers across our network and further stratify these performances across skin tone in Figure \ref{fig:ss_plot_ita}.

\begin{table}[!t]
    \centering
    \begin{tabular}{c|cc}
    \toprule Methods &AUROC& $F_1$\\ \midrule
    \multirow{1}{*}{Softmax Score~\citep{hendrycks}} & $74.4\pm1.7$  & ${71.0\pm1.1}$  \\
    \multirow{1}{*}{ODIN~\citep{liang}} &  ${74.5\pm1.6}$ & $70.8\pm1.1$ \\
    \midrule
    \multirow{1}{*}{SS ($C_s$)} & $68.2\pm1.4$ & $71.3\pm0.5$ \\
    \multirow{1}{*}{SS ($C_{gp}$)} & $62.7\pm1.2$ & ${72.5\pm0.6}$ \\
    \multirow{1}{*}{SS ($C_{conv_1}$)} & $41.6\pm1.8$ & $68.1\pm0.2$ \\
    \midrule
    \multirow{1}{*}{SS ($C_s$)+ODIN } & $51.2\pm1.9$ & $67.9\pm0.3$ \\
    % \multirow{1}{*}{SS (Sum All Layers)+ODIN}& $9.9\pm1.0$  & $67.6\pm0.03$\\
    \multirow{1}{*}{SS ($C_{conv_1}$)+ODIN$_{low}$} & $85.4\pm0.6$ & $81.9\pm0.6$\\
    \multirow{1}{*}{SS (Sum All Layers)+ODIN$_{low}$} & $\mathbf{{91.0\pm0.8}}$ & $\mathbf{{86.9\pm1.1}}$\\
    % \midrule
    \bottomrule
    \end{tabular}
    \caption{Detection performance for OOD samples of unknown collection protocols validated with SD-198~\citep{sd-198}. Bold values are the best performers in each column.}
    \label{tab:lr_bin_sd}
    %\vspace{-3mm}
\end{table}

\subsection{OOD samples from a different protocol or equipment}
We first show the result of detecting OOD samples that are collected with different protocols or equipment.
% than the training samples of the network (ISIC 2019). 
Table~\ref{tab:lr_bin_sd} summarizes the results of the proposed approach - subset scanning (SS) with and without noise, and compared with the existing baselines~\citep{hendrycks,liang}. 
% Bold values show the best performance in the entire column. 
In the top panel, we see that ODIN~\citep{liang} increases the AUROC performance of Softmax Score by around 0.1 on average.
% As ODIN is optimized over the SoftMax Score, we show the result of applying subset scanning on SoftMax layer $C_{s}$ for samples perturbed by ODIN. 
For samples with ODIN noise, we show the performance of subset scanning on the softmax layer $C_{s}$, as ODIN is optimized on Softmax Score, and for samples with ODIN$_{low}$ noise, we show the result of subset scanning on the first convolutional layer ($C_{conv_1}$). We achieve the best performance with AUROC of $91.0 \pm 0.8$ and maximum $F_1$-score of $86.9\pm1.1$ using the sum of subset scores $S^*_{(C_y)}$ across all eight layers with ODIN$_{low}$ (bottom row in Table~\ref{tab:lr_bin_sd}).

\setlength{\tabcolsep}{5pt}
\begin{table}[!h]
    \centering
    % \resizebox{1\linewidth}{!}{
    \begin{tabular}{c|cc|cc}
    \toprule \multirow{2}{*}{Methods}& \multicolumn{2}{c}{AUROC}& \multicolumn{2}{|c}{ $F_1$}\\
    & DF & VASC  & DF & VASC \\ \midrule
    \multirow{1}{*}{Softmax Score~\citep{hendrycks}} & \textbf{80.9} & \textbf{73.2} & \textbf{76.5} & 70.5  \\
    \multirow{1}{*}{ODIN~\citep{liang}} & 72.3 & 65.3  & 70.3 & 67.4 \\ 
    \midrule
    \multirow{1}{*}{SS ($C_s$)} & 80.8 & 70.8 &75.7 & \textbf{72.3} \\
    \multirow{1}{*}{SS ($C_{gp}$)} & 37.4 & 57.9 &65.9 & 69.2 \\
    \multirow{1}{*}{SS ($C_{conv_1}$)} & 50.9 & 62.5 &65.8 &68.7 \\ \midrule
    \multirow{1}{*}{SS ($C_s$)+ODIN} & 71.8 & 63.3 & 70.4 & 67.4 \\
    \multirow{1}{*}{SS ($C_{conv_1}$)+ODIN$_{low}$} & 47.6 & 39.8 & 65.9 & 67.1 \\
    \multirow{1}{*}{SS (Sum All Layers)+ODIN$_{low}$} & 47.6 & 40.4 & 65.9 & 67.2 \\
    \bottomrule
    \end{tabular}
    % }
    \caption{Performances of detecting OOD samples of unknown disease types, DF and VASC. Bold values are the best performers in each column.}
    \label{tab:lr_bin_isic}
    %\vspace{-3mm}
\end{table}

\subsection{OOD samples of unknown diseases}
Table~\ref{tab:lr_bin_isic} shows the performance of detecting OOD samples of unknown diseases (DF and VASC) that are unseen during training. While Softmax Score~\citep{hendrycks} yields the best performance, subset scanning on the softmax layer $C_s$ shows comparable performance. We see worse performances with ODIN as these OOD samples are from the same dataset as ID samples and adding noise likely blurs the unique features present in each skin disease class.

\setlength{\tabcolsep}{5pt}
\begin{table*}[!h]
    \centering
    % \resizebox{1\linewidth}{!}{
    \begin{tabular}{cc|cc|cc|cc}
    \toprule \multirow{3}{*}{Methods} &\multirow{3}{*}{Skin Tone}&  \multicolumn{4}{c|}{Unknown diseases}&  \multicolumn{2}{c}{Collection protocol}\\ \cline{3-8}
    & & \multicolumn{2}{c|}{DF}& \multicolumn{2}{c|}{VASC} & \multicolumn{2}{c}{SD-198}\\ 
    & & R & AUROC & R & AUROC & R & AUROC \\ \midrule \midrule
    \multirow{3}{*}{\begin{tabular}{c} Softmax Score~\citep{hendrycks}\end{tabular}}  
     & Light  & 171 & \textbf{81.0}& 185 &72.1 & 986 & \textbf{75.8}\\
     & Intermediate & 52 &80.7 & 58 & 75.8 &1278 & 73.7\\
     & Dark  & 10 & 74.9& 9 & \textbf{77.0} & 326 &73.2\\ \midrule
    \multirow{3}{*}{\begin{tabular}{c}ODIN~\citep{liang}\end{tabular}}  
     & Light  & 171 & 71.6& 185 & 64.0 & 986 & \textbf{76.2} \\
     & Intermediate & 52 & 69.9& 58 & 64.9 &1278 & 73.8\\
     & Dark  & 10 & \textbf{86.3} & 9 &\textbf{89.4} &326 & 72.1\\
     \midrule \midrule
    \multirow{3}{*}{SS ($C_s$)}  
     & Light  & 171 & 78.6& 185 & 70.7 & 986 & 68.3\\
     & Intermediate & 52 & 87.0& 58 &\textbf{71.3} & 1278 & 68.0\\
     & Dark  & 10 & \textbf{87.6} & 9 & 69.5 & 326 & \textbf{68.6}\\ % \midrule
    % \multirow{3}{*}{SS ($C_{conv_1}$)}  
    %  & Light  & 171 & 47.9& 185 & 60.2 & 986 & 42.9\\
    %  & Intermediate & 52 & 50.8 & 58 & 67.9 & 1278 & 38.2\\
    %  & Dark & 10 & \textbf{89.2} & 9 & \textbf{70.7} & 326 & \textbf{50.5}\\ 
    \midrule \midrule
    \multirow{3}{*}{SS ($C_s$)+ODIN}  
     & Light  & 171 & 69.7 & 185 & 62.7 &986 & \textbf{52.1}\\
     & Intermediate & 52 & 73.8 & 58 & 63.1 &1278 & 50.6\\
     & Dark  & 10 & \textbf{88.2} & 9 & \textbf{74.5} &326 & 50.9\\
     \midrule
    \multirow{3}{*}{\begin{tabular}{c}SS ($C_{conv_1}$) \\ + ODIN$_{low}$ \end{tabular}} 
     & Light  & 171 & 45.1& 185 & 38.8 & 986 &83.1\\
     & Intermediate & 52 & 49.9 & 58 & 37.8 &1278 & 86.7\\
     & Dark  & 10 & \textbf{63.6}& 9 &\textbf{68.4} &326 & \textbf{87.2}\\ 
     \midrule
    \multirow{3}{*}{\begin{tabular}{c}SS (Sum All Layers) \\ + ODIN$_{low}$\end{tabular}}  
     & Light  & 171 & 45.1& 185 & 38.4 & 986 & 89.3\\
     & Intermediate & 52 &51.8 & 58 & 40.0 & 1278 & 92.0\\
     & Dark  & 10 & \textbf{56.2}& 9 &\textbf{78.7} & 326 & \textbf{92.3}\\  
     \bottomrule
    \end{tabular}
    % }
    \caption{Performance of methods in Tables~\ref{tab:lr_bin_sd} and \ref{tab:lr_bin_isic} stratified into three different skin tone categories. $R$ represents the number of OOD samples in each category.}\label{tab:lr_ita_bin}
    %\vspace{-7mm}
\end{table*}

\subsection{Performance stratified by skin-tone} 
% \hannah{(Side Comment: The sample number $N$ do not sum up to 260*10=2600 because 10 samples came out as undefined by Kinyanjui \textit{et al.}~\cite{skinTone}.)}
We further stratify the OOD samples into three skin tone categories and show the results in Table~\ref{tab:lr_ita_bin}. In each set of columns, we include the number of test samples $R$ for each skin tone category and its corresponding AUROC performance. Samples of Dark skin tones constitute only around $3.9\%$ of DF and VASC samples and around $13\%$ of SD-198 samples.
%\girmaw{
Majority of the listed methods (13 out of 18), show higher detection performance of Dark OOD samples. This could be partially because network is trained on the ISIC 2019 dataset that heavily lacks samples of dark skin tones, and thus easily detects OOD samples of dark skin tone to be out of distribution. Overall, it requires further investigation to clearly understand whether such performance reveals the lack of Dark samples in these datasets or variant manifestations of skin diseases in Dark skin.
% Looking at the best performing skin tone in each category (bolded), we see that out of the 18 methods we consider, 13 of them show the best performance with samples of Dark skin tone, 4 are from those of Light skin tone, and only 1 is from Intermediate skin tone. This is expected as our network is trained on the ISIC 2019 dataset that heavily lacks samples of dark skin tones, and thus easily detects OOD samples of dark skin tone to be out of distribution. 
% \vspace{-0.6cm}

\setlength{\tabcolsep}{0pt}
\begin{figure}[!h]
    \centering
    % \resizebox{\linewidth}{!}{
    \begin{tabular}{ccc}
    \toprule
        SS & SS+ODIN &SS+ODIN$_{low}$ \\\midrule
        \includegraphics[width = 0.33\linewidth]{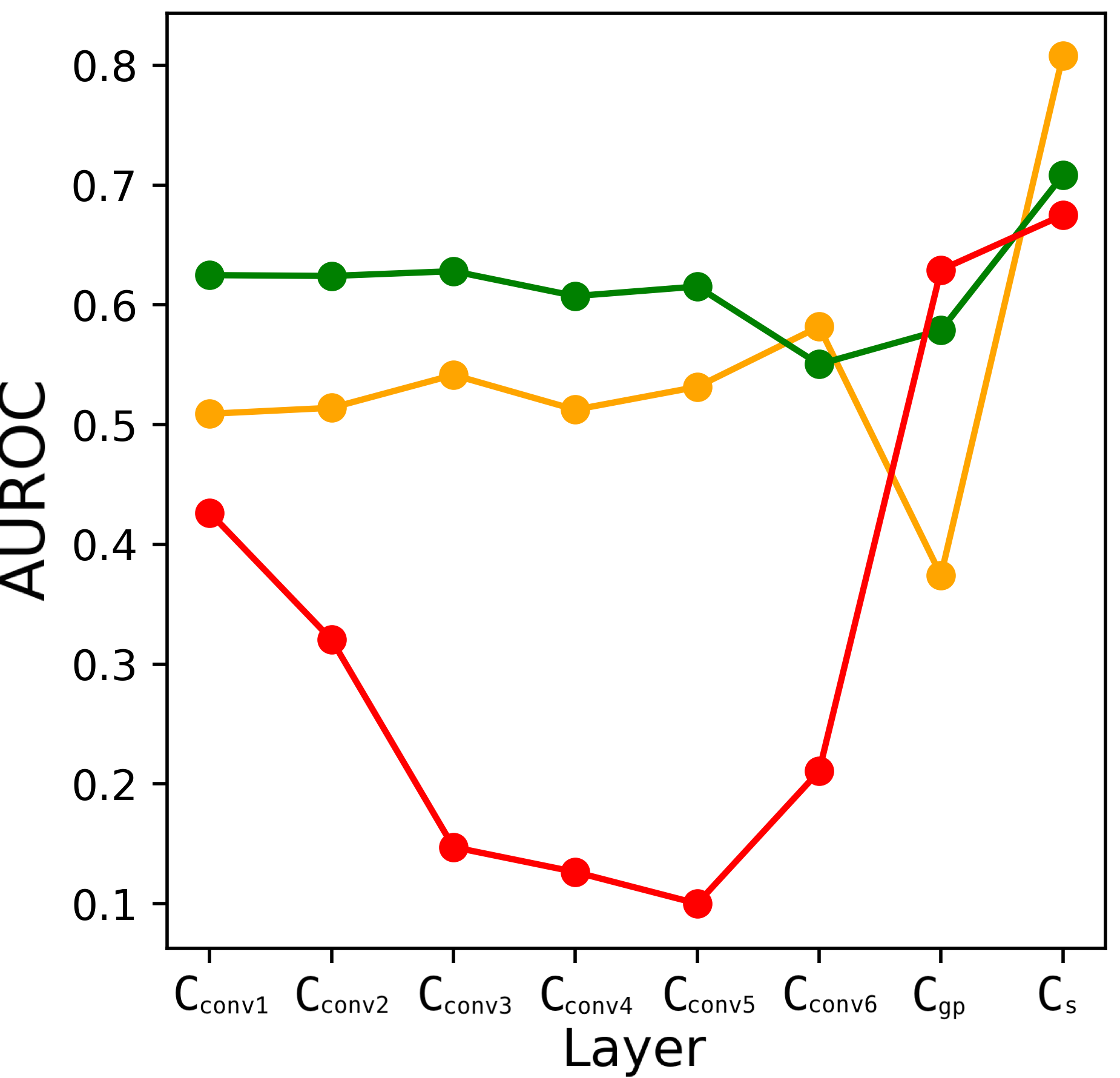} &
        \includegraphics[width = 0.33\linewidth]{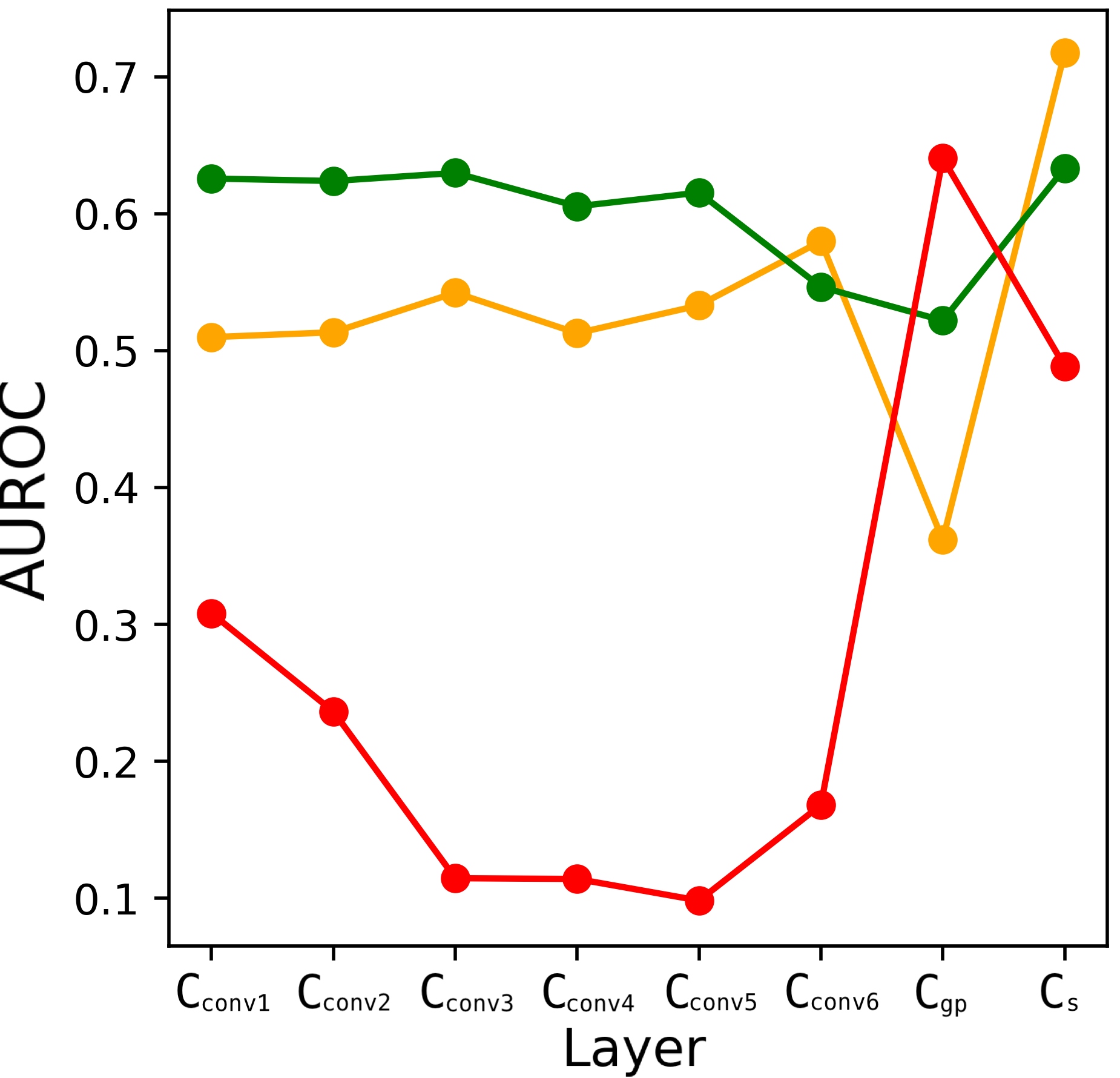} &
         \includegraphics[width = 0.33\linewidth]{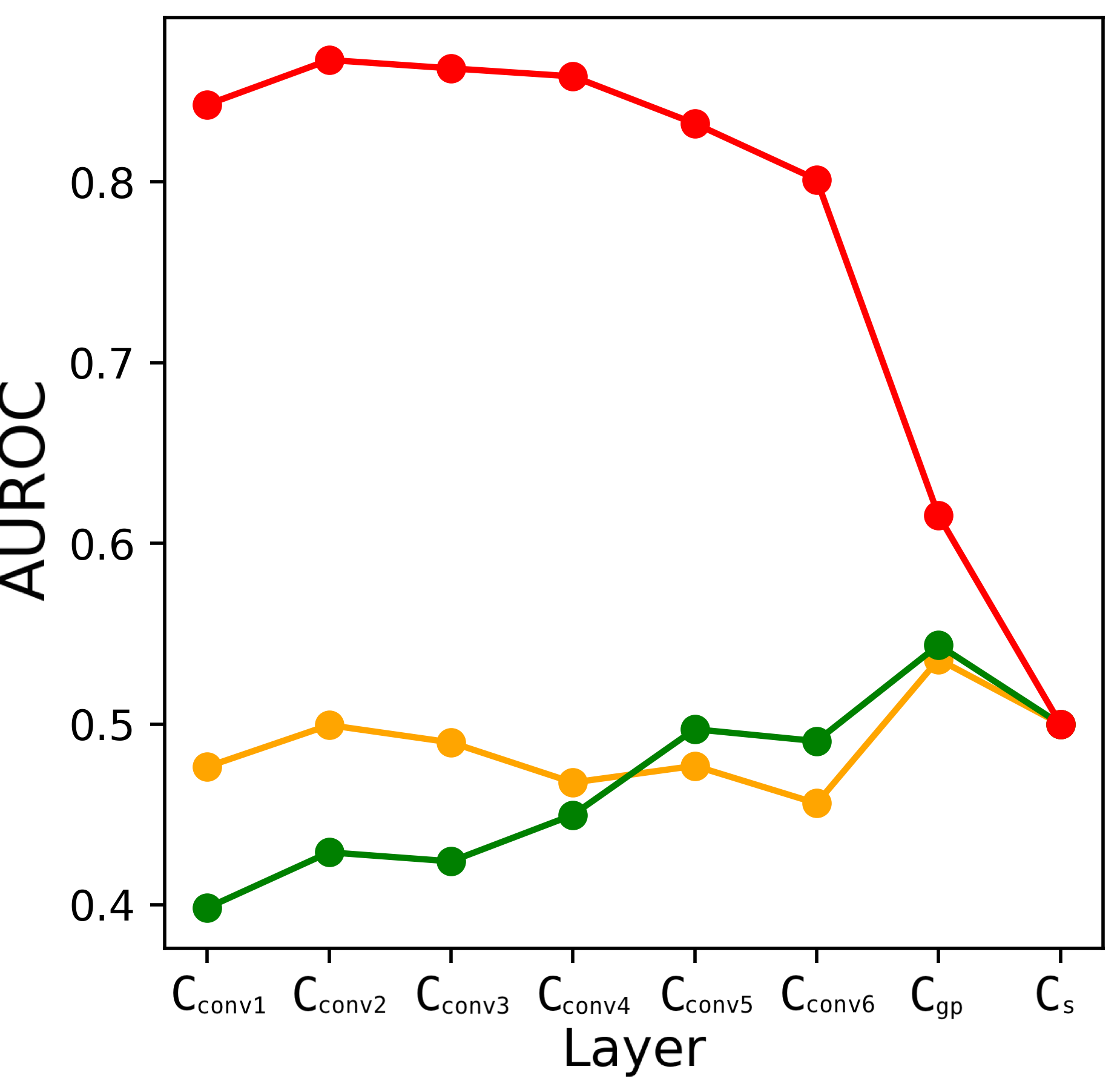} \\
         \bottomrule
    \end{tabular}
    % }
    \caption{AUROC performance of subset scanning (SS) across various layer of DenseNet-121 that we consider. First column shows the results without any ODIN, the other two columns show the result with ODIN and ODIN$_{low}$, respectively for OOD samples of DF (yellow), VASC (green), and SD-198 (red).
    % \textcolor{red}{@Hannah - do we want to change sd-196 in the legends to SD-198?}
    }
    \label{fig:ss_plot}
\end{figure}

\subsection{OOD detection across individual layers}
Figure~\ref{fig:ss_plot} shows the OOD detection performance in terms of AUROC of our proposed work on the eight layers of our pre-trained CNN in $C_Y$ that we consider. The first column shows the result of subset scanning without any added noise, and the other columns show the result of applying ODIN~\citep{liang} and ODIN$_{low}$ perturbations, respectively, to our test images before applying subset scanning. In each sub-plot, we show results of both use cases, i.e., detection of samples of unknown diseases (DF (yellow), VASC (green)) and samples from different protocols (SD-198 (red)). %\hannah{Not entirely sure of the conclusion we want to make from this figure.} 
Overall, DF and VASC samples from ISIC 2019 dataset have similar performance across the eight layers we consider while samples from SD-198 dataset leads to varying performances depending on the layer and ODIN parameters. This is partly because DF and VASC samples are from the same distribution as the training set as they are both from the same ISIC 2019 dataset, while SD-198 has different distribution than the training set of ISIC 2019 with different collection protocol. Comparing the last two plots, we see that standard ODIN leads to better performance near the end of the network while ODIN$_{low}$ leads to better performance in earlier layers of the network. This is as expected as ODIN parameters ($\tau$ and $\epsilon$) are optimized on the Softmax  Scores while ODIN$_{low}$ parameters, $\tau_{low}$ and $\epsilon_{low}$, are not.

\setlength{\tabcolsep}{0pt}
\begin{figure}[htbp!]
    \centering
    % \resizebox{\linewidth}{!}{
    \begin{tabular}{ccc}
    \toprule
        SS & SS+ODIN & SS+ODIN$_{low}$\\
        %  & $T=5$, $\epsilon=0.0002$ & $T=2$, $\epsilon=0.2$ \\
        \midrule
        \includegraphics[width = 0.33\linewidth]{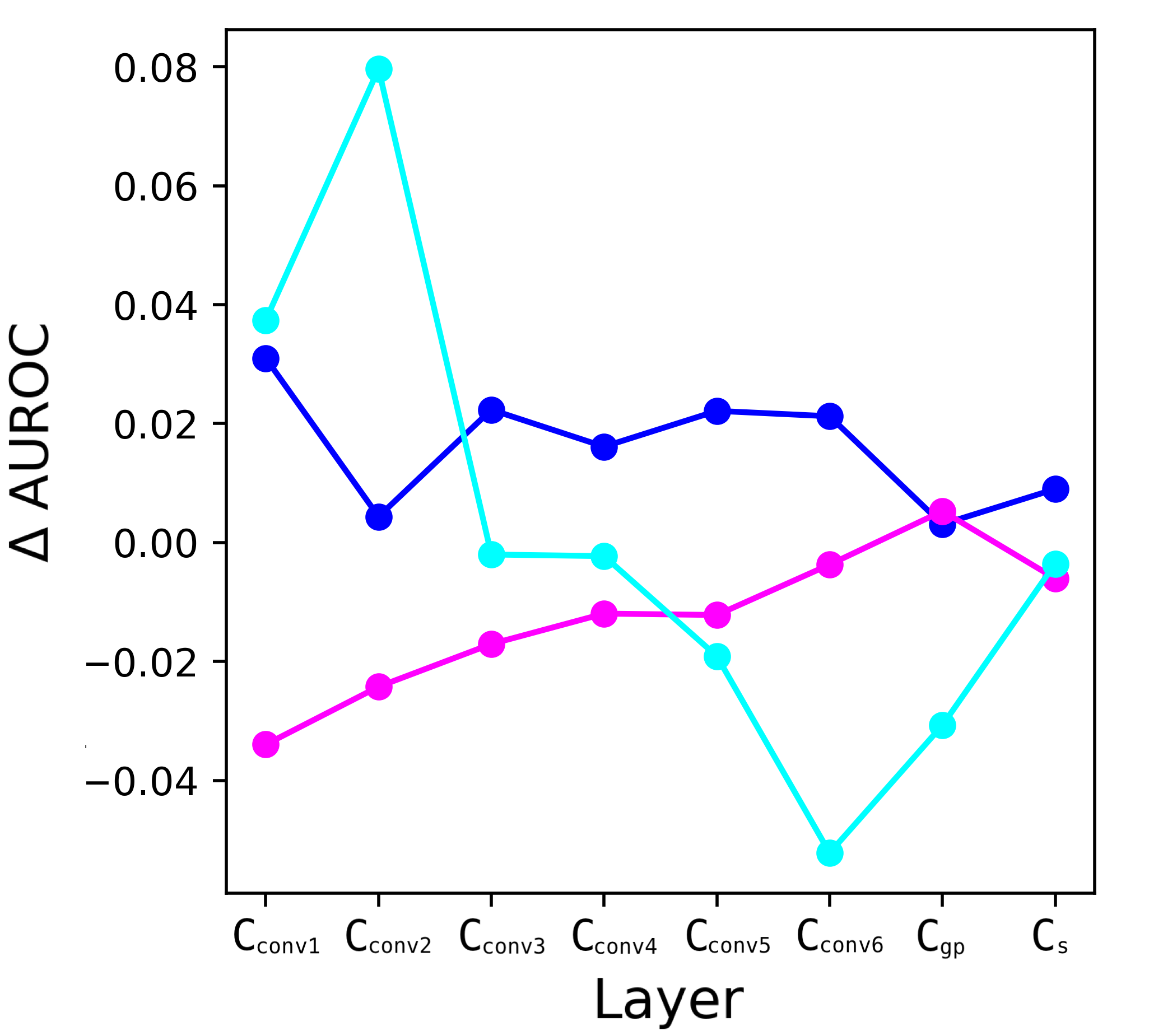} &
        \includegraphics[width = 0.33\linewidth]{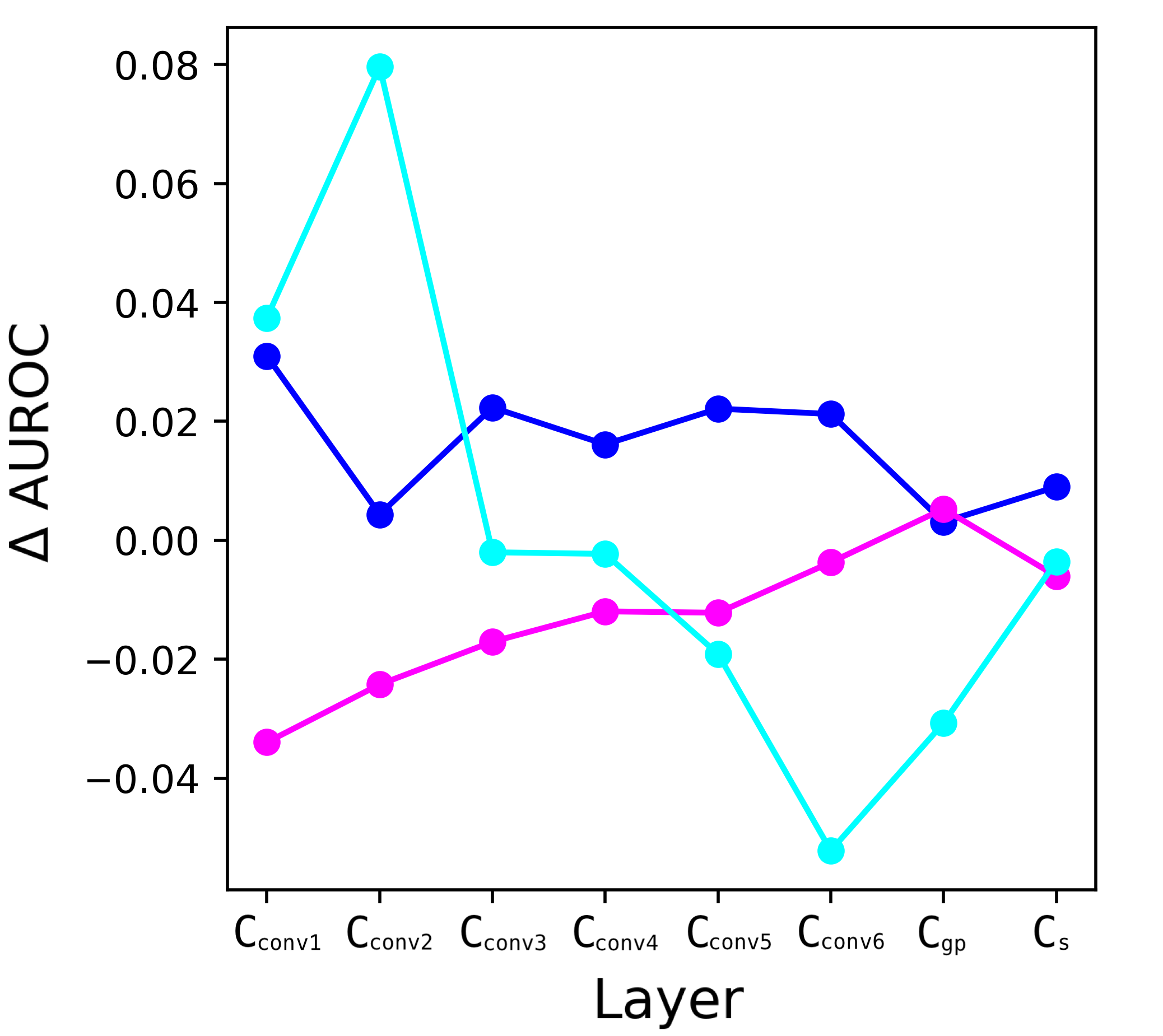} &
         \includegraphics[width = 0.33\linewidth]{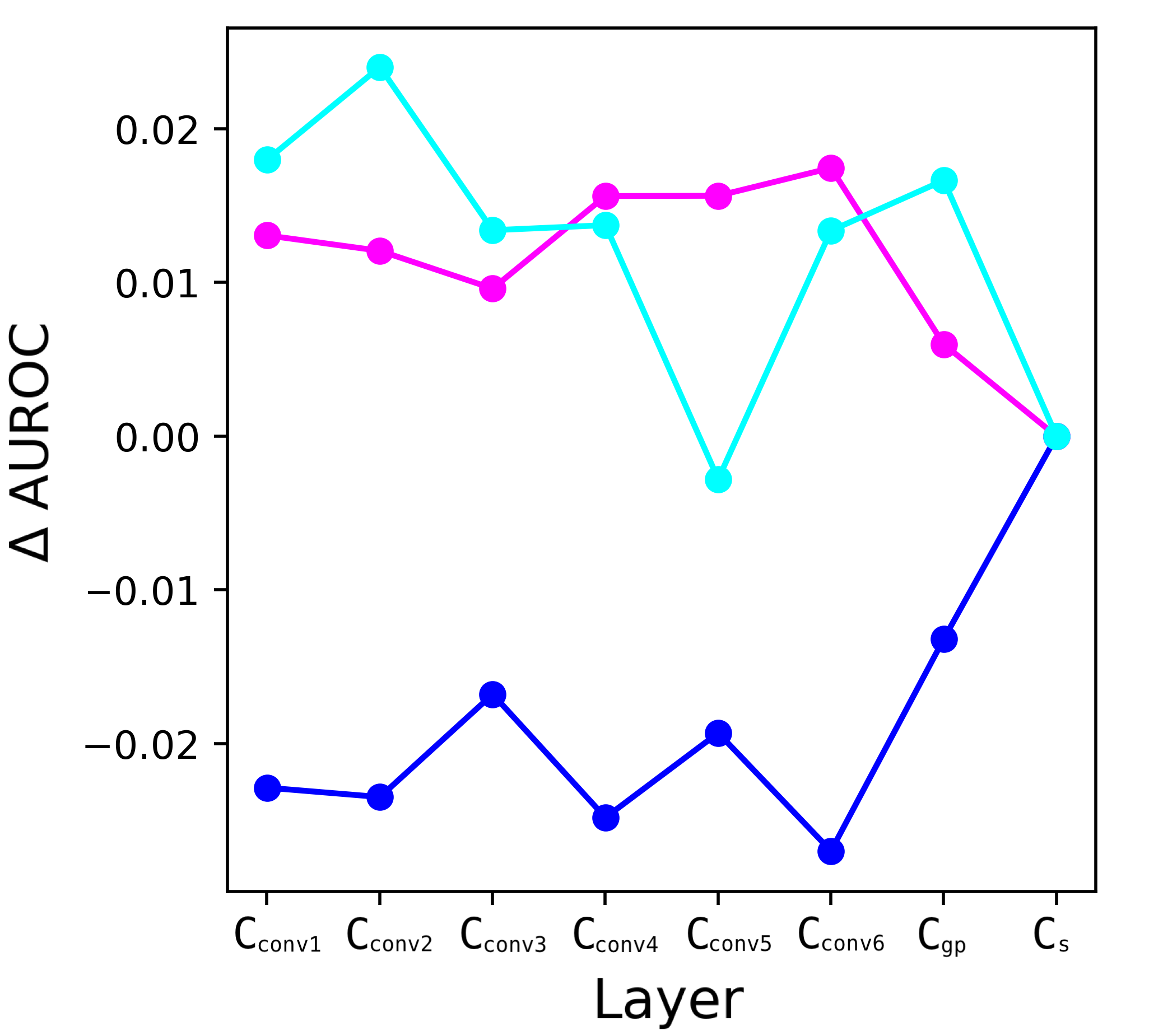} \\
        % \includegraphics[width = 0.3\textwidth]{images/ss_individual_auroc_sd_ita2.png} &
        % \includegraphics[width = 0.3\textwidth]{images/ss_individual_auroc_sd_perturb_t10_m_ita2.png} &
        %  \includegraphics[width = 0.3\textwidth]{images/ss_individual_auroc_sd_perturb_t2_m2_ita2.png} \\
        % \multicolumn{3}{c}{SD-198}\\ 
        \midrule
        \includegraphics[width = 0.33\linewidth]{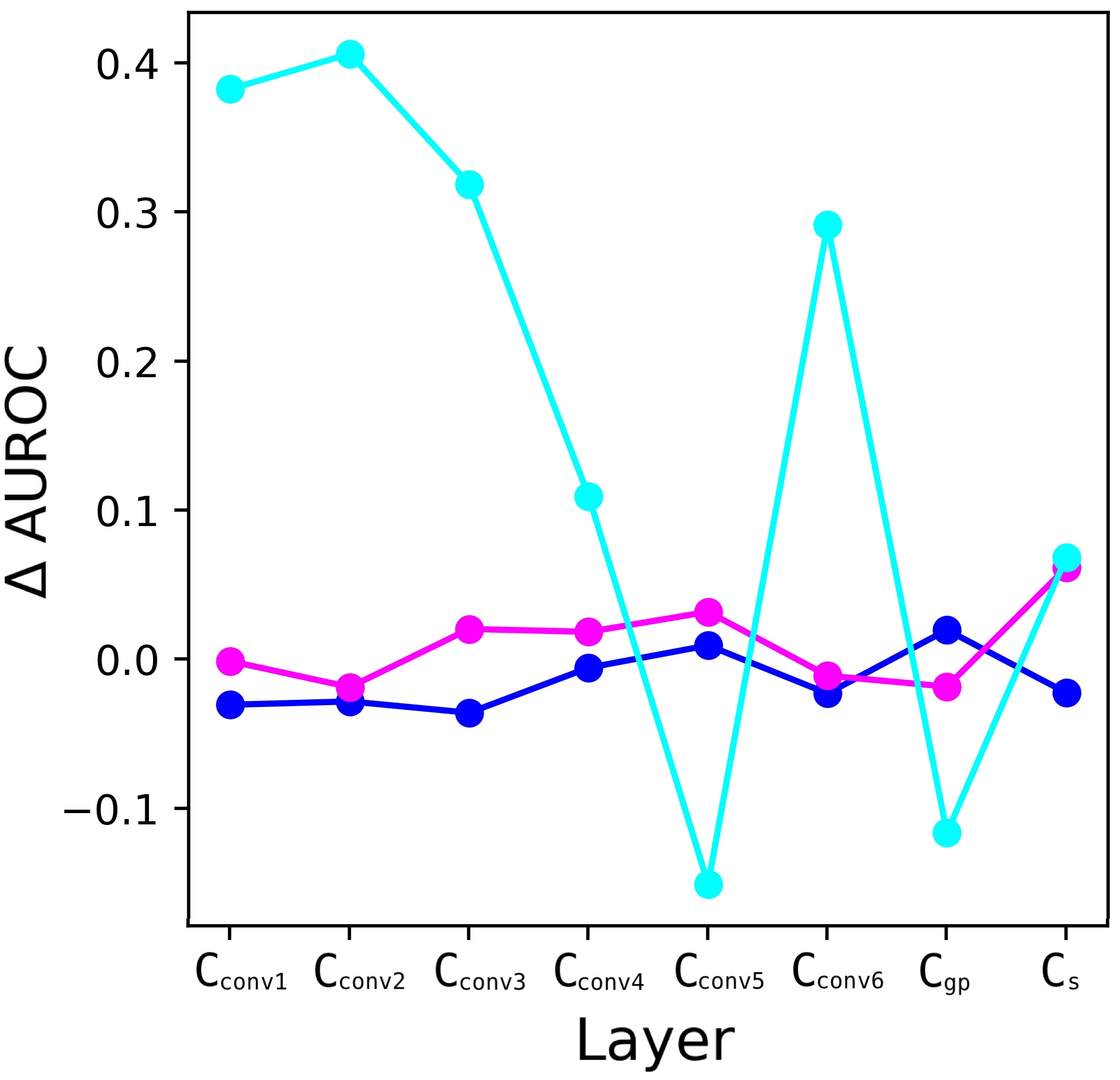} &
        \includegraphics[width = 0.33\linewidth]{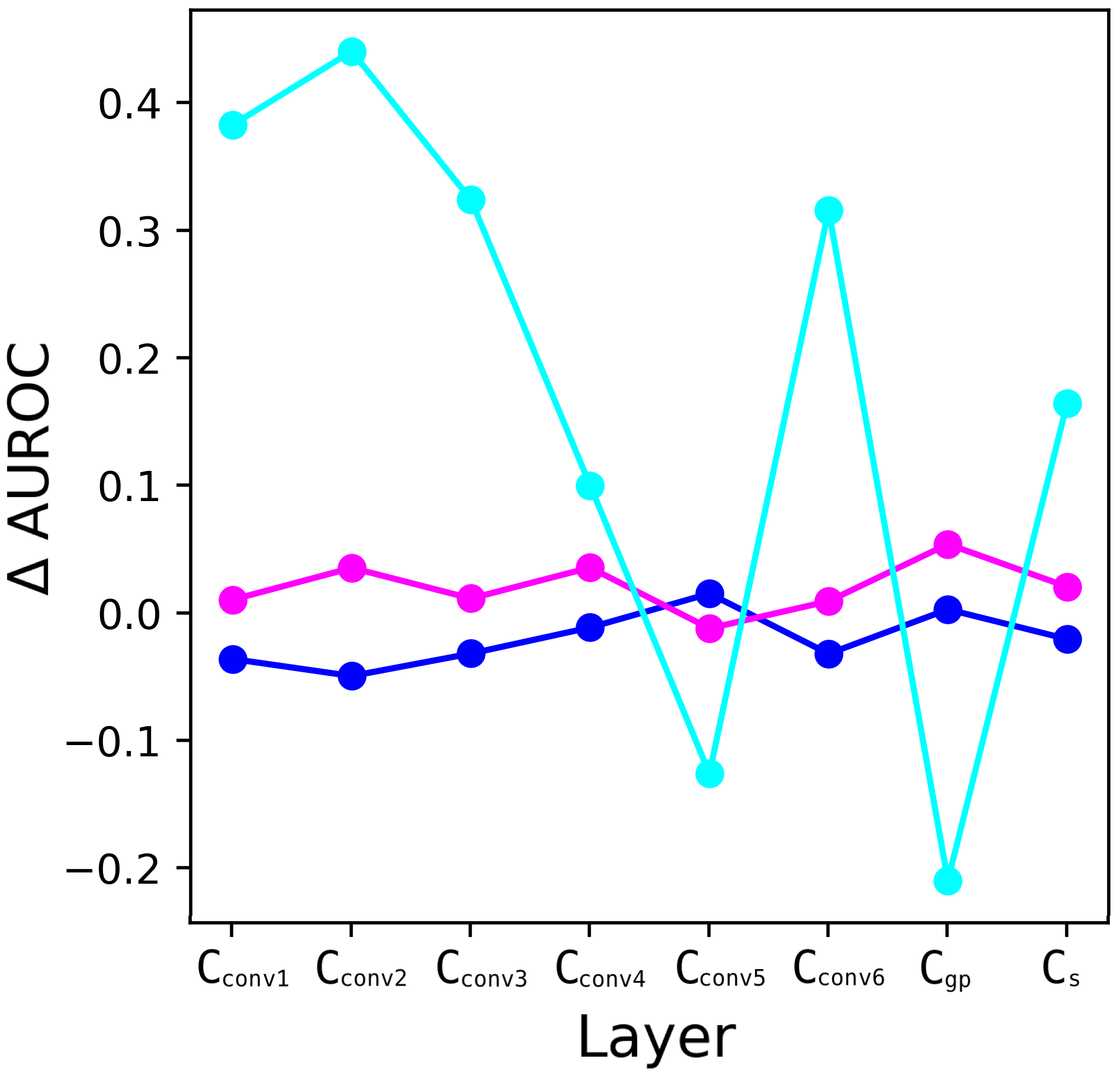} &
         \includegraphics[width = 0.33\linewidth]{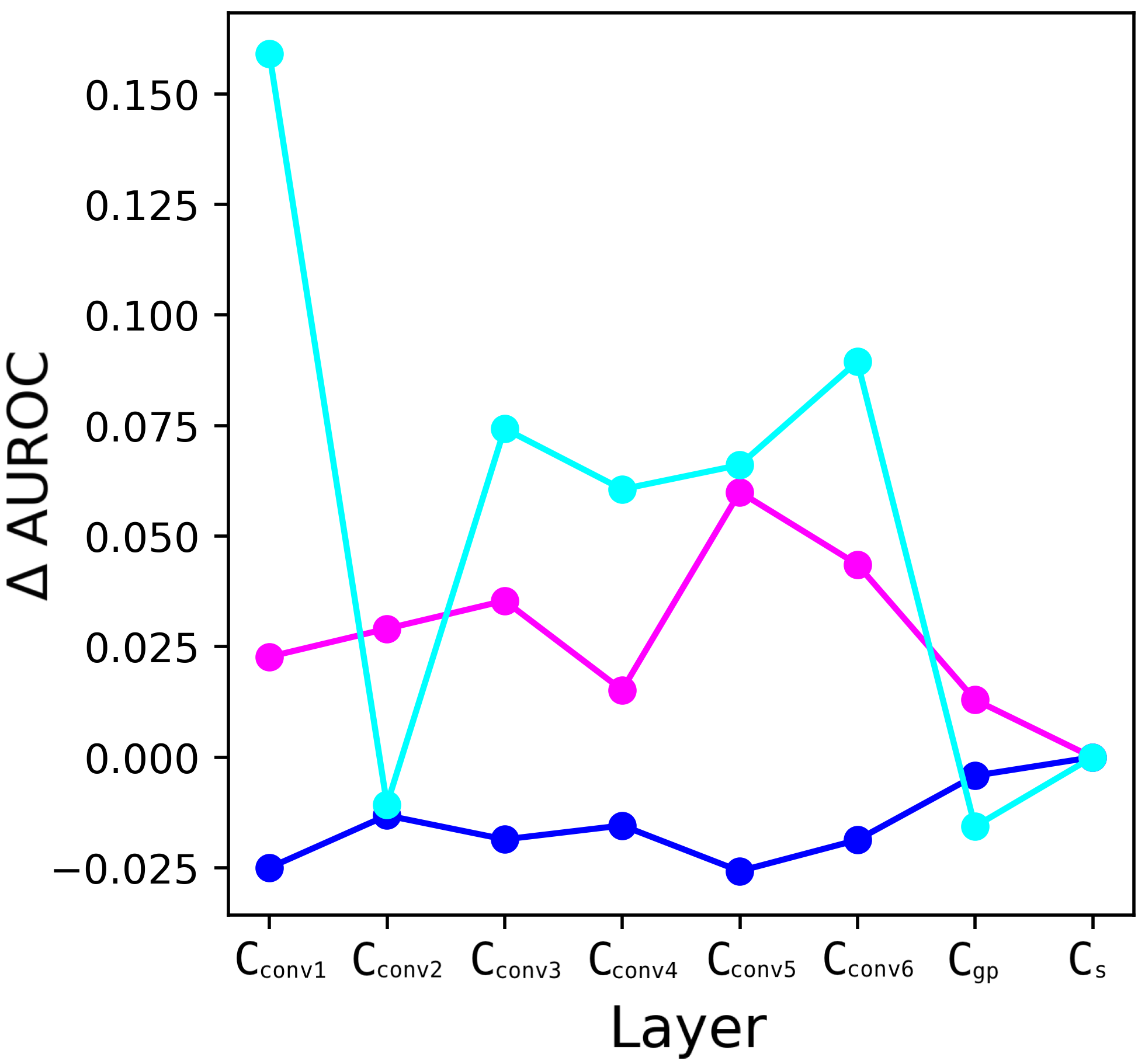} \\
        % \includegraphics[width = 0.2\textwidth]{images/ss_individual_auroc_DF_diff_perturb_t1_m002_ita2.png} \\
        % \multicolumn{3}{c}{DF}\\ 
        \midrule 
        \includegraphics[width = 0.33\linewidth]{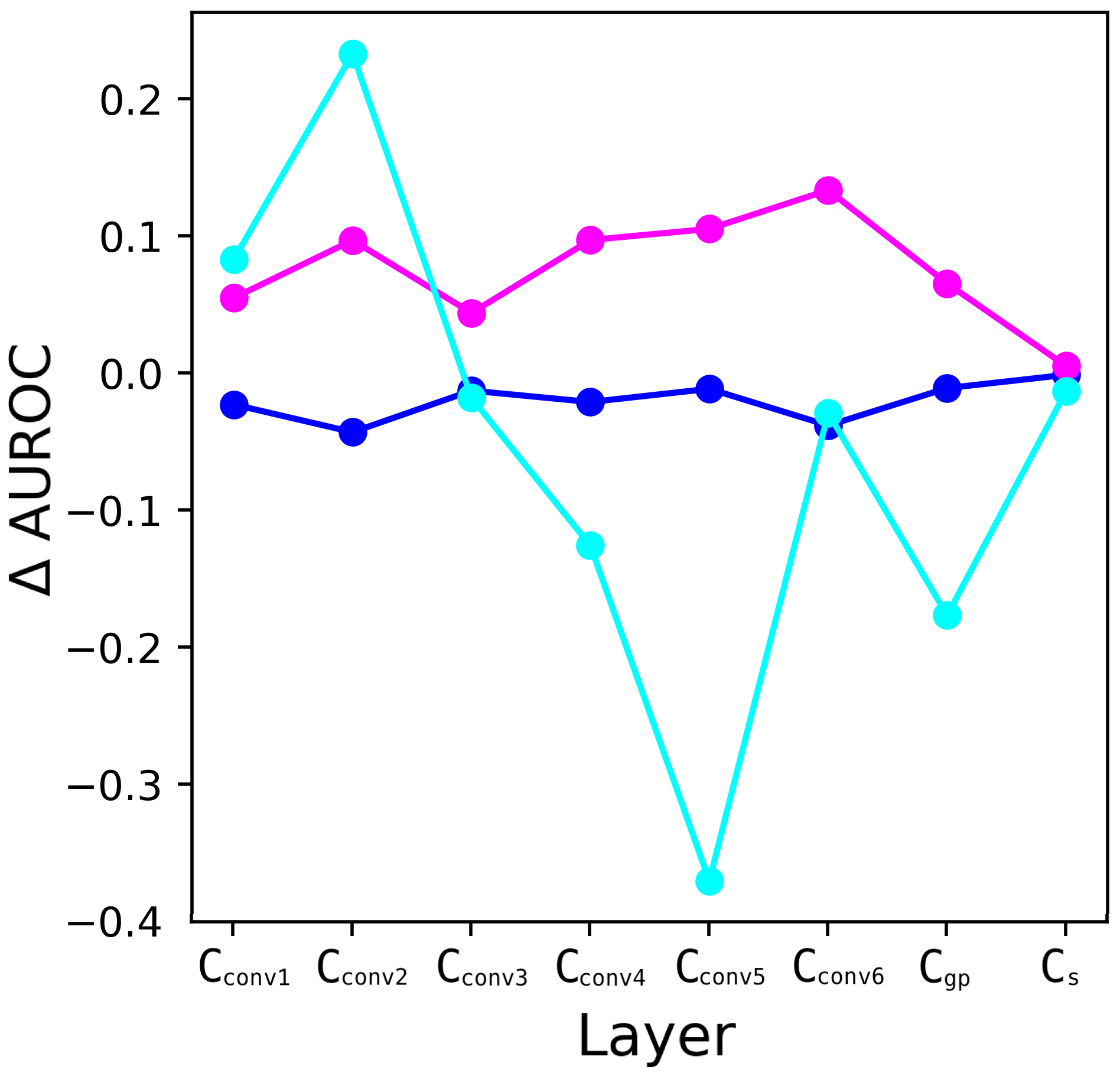} &
        \includegraphics[width = 0.33\linewidth]{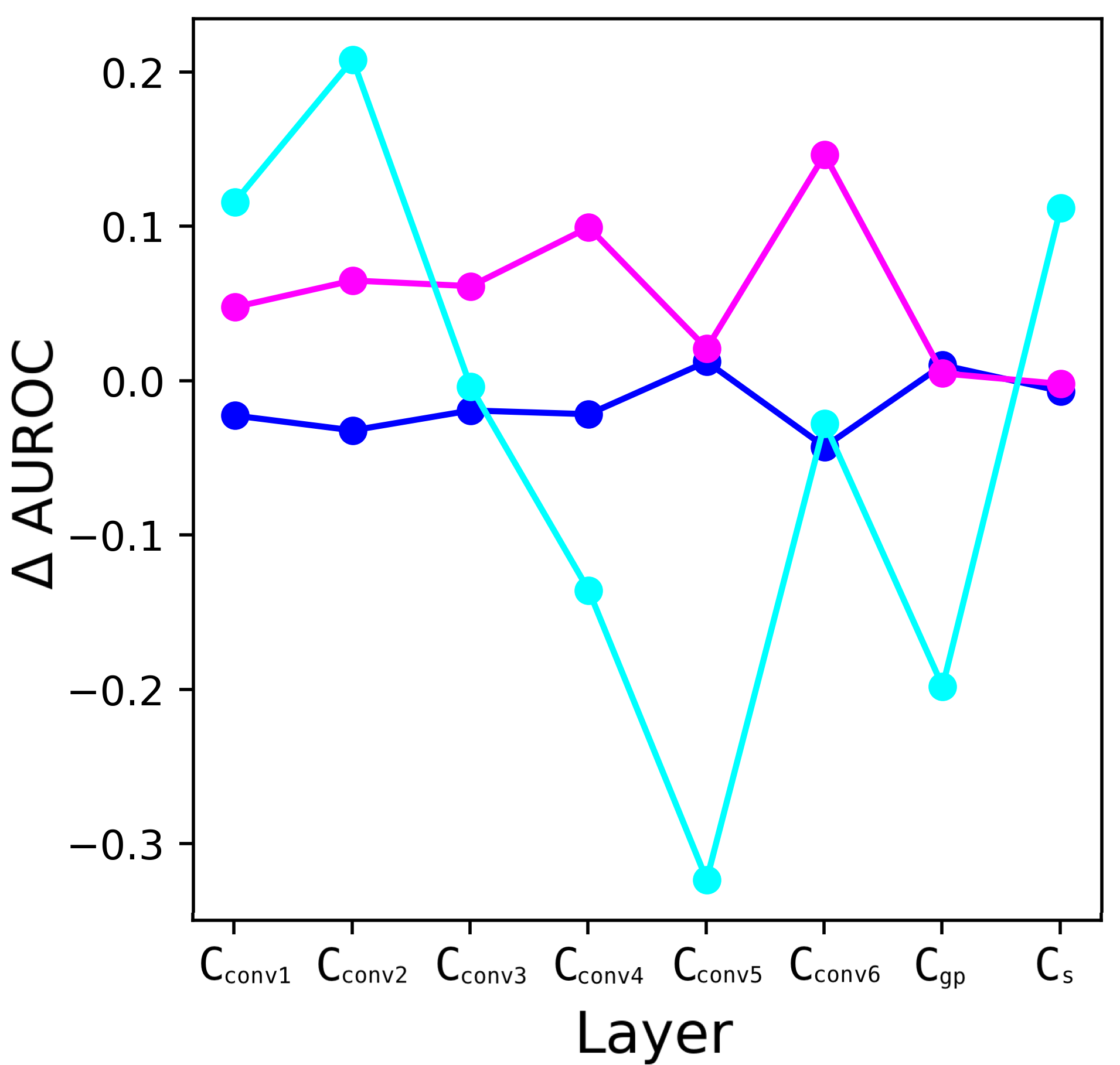} &
         \includegraphics[width = 0.33\linewidth]{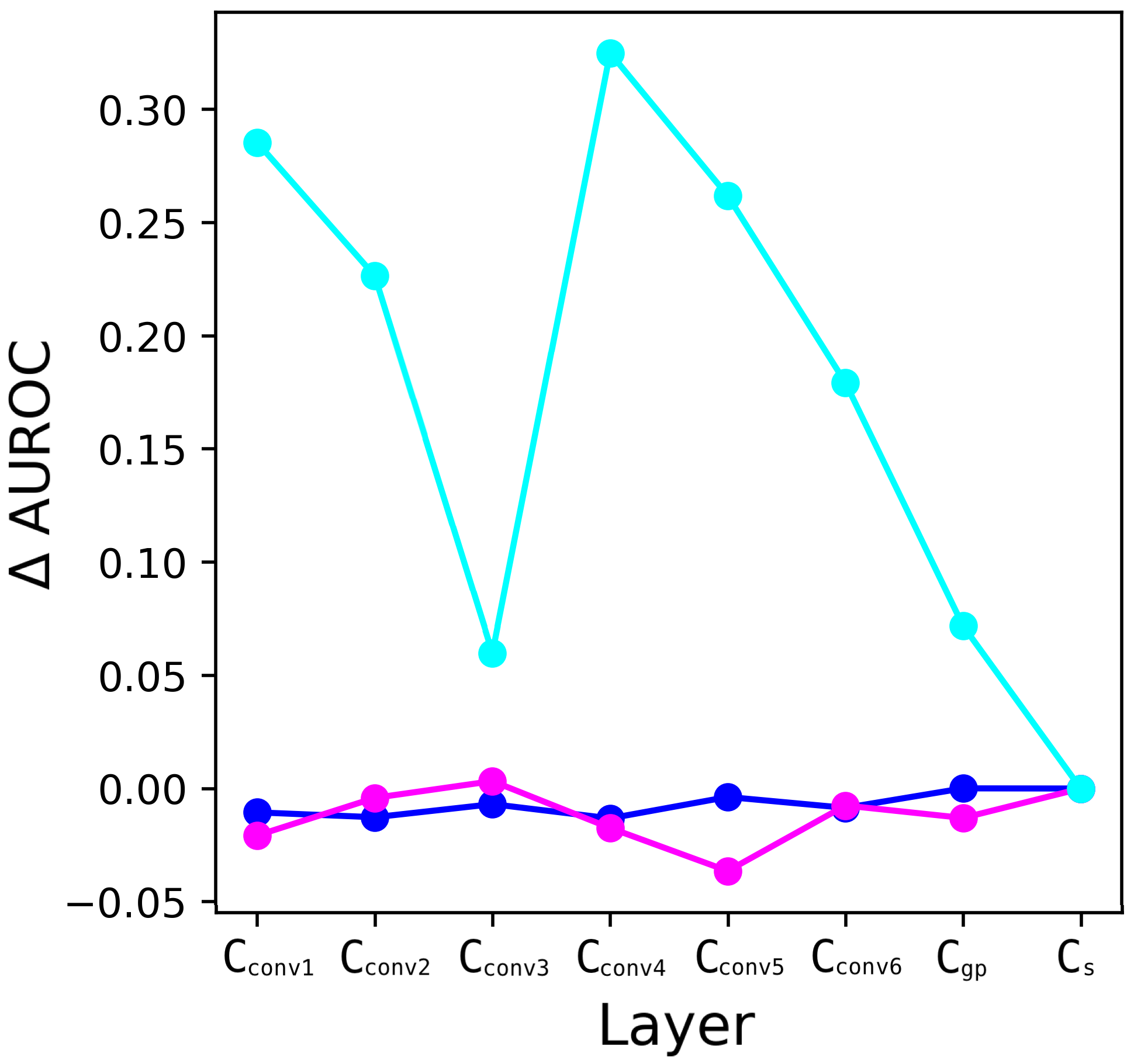} \\
        % \includegraphics[width = 0.2\textwidth]{images/ss_individual_auroc_VASC_diff_perturb_t1_m002_ita2.png}  \\
        % \multicolumn{3}{c}{VASC}\\ 
        \bottomrule
    \end{tabular}
    % }
    \caption{Change in performance ($\Delta$ AUROC) of OOD detection in Figure \ref{fig:ss_plot} for SD-198 (top), DF (middle) and VASC (bottom) stratified into three different skin-tone categories, Light (blue), Intermediate (magenta), and Dark (cyan). First column shows the results without any ODIN, the other two columns show the result with ODIN and ODIN$_{low}$, respectively.}
    \label{fig:ss_plot_ita}
\end{figure}

We further stratify the performance of individual layers based on skin tone represented in the samples and show the change in AUROC with the stratification in Figure~\ref{fig:ss_plot_ita}. While the samples of Light (blue) and Intermediate (magenta) skin tones show consistent performances throughout the layers, we see varying performances for samples of Dark (cyan) skin tones. This instability of performance for samples of Dark skin tones may be partially because network is trained on the ISIC 2019 dataset that heavily lacks samples of Dark skin tones.

\section{Conclusion}\label{conclusion}
% \vspace{-2mm}
We propose a weakly-supervised method to detect out-of-distribution  (OOD) skin images (collected in different protocols or from unknown disease types) using input perturbation and scanning of the activations in the intermediate layers of pre-trained on-the-shelf classifier. The scanning of activations is optimised as a search problem to identify nodes in a layer that results in maximum divergence of the activations from subset of test samples compared to the expected activations derived from the training (in-distribution) samples.  We exploited Linear Time Subset Scanning (LTSS)~\citep{neill-ltss-2012} property of subset scanning to achieve efficient search that scales linearly with the number of nodes in the a layer.  Our proposed method improves on the state-of-the-art detection for OOD samples that are collected from a different protocol or equipment than those in-distribution samples used to train the classifier, and it achieves competitive performance with the state-of-the-art in detecting samples of unknown diseases. 
% For OOD samples that are from the same dataset or distributions the training dataset (ISIC 2019 dataset) but are from a new unknown class labels (DF and VASC), we do comparable with the existing work. 
We further stratify these OOD samples based on three skin tone categories, Light, Intermediate, and Dark. %\girmaw{
From our results we observe imbalanced detection performance across skin tones, where the Dark samples are detected as OOD with higher performance. Thus, future work aims to understand the reasons for such detection disparity across skin tones, e.g., lack of training representation or different manifestation of skin diseases.

\bibliographystyle{unsrtnat}
\bibliography{main.bib}

\begin{thebibliography}{35}
\bibitem[rab()]{rabin_2020}
Dermatology has a problem with skin color.
\end{thebibliography}
\appendix

%\section{Cardinality and Alpha values distribution}

\end{document}